# Exploring Artificial Intelligence Methods for Energy Prediction in Healthcare Facilities: An In-Depth Extended Systematic Review

Marjan FatehiJananloo, Helen Stopps, J.J. McArthur*

Dept. Architectural Science, Toronto Metropolitan University

*corresponding author: jjmcarthur@torontomu.ca

**Abstract:**

Hospitals, due to their complexity and unique requirements, play a pivotal role in global energy consumption patterns. This study conducted a comprehensive literature review, utilizing the PRISMA framework, of articles that employed machine learning and artificial intelligence techniques for predicting energy consumption in hospital buildings. Of the 1884 publications identified, 17 were found to address this specific domain and have been thoroughly reviewed to establish the state-of-the-art and identify gaps where future research is needed. This review revealed a diverse range of data inputs influencing energy prediction, with occupancy and meteorological data emerging as significant predictors. However, many studies failed to delve deep into the implications of their data choices, and gaps were evident regarding the understanding of time dynamics, operational status, and preprocessing methods. Machine learning, especially deep learning models like ANNs, have shown potential in this domain, yet they come with challenges, including interpretability and computational demands. The findings underscore the immense potential of AI in optimizing hospital energy consumption but also highlight the need for more comprehensive and granular research. Key areas for future research include the optimization of ANN approaches, new optimization and data integration techniques, the integration of real-time data into Intelligent Energy Management Systems, and increasing focus on long-term energy forecasting.

# 1   Introduction

Hospitals are complex structures with unique requirements, distinguishing them from other large buildings. These facilities demand specialized considerations, including state-of-the-art medical equipment, stringent hygiene protocols, and ventilation systems (Xue, et al., 2020). These extensive needs of hospitals mandate the continuous operation of complex heating, ventilation, and air conditioning (HVAC) systems, which account for a substantial portion of all building energy consumption (Coccagna, et al., 2017). A study conducted in England revealed that hospital building services contributed to a staggering 40% of the total national greenhouse gas emissions within the hospital and related services sector (Panagiotou & Dounis, 2022). As a result, hospitals have emerged as a critical area of focus for enhancing energy efficiency, offering substantial potential to reduce carbon emissions and contribute to sustainable development goals. Consequently, prioritizing the optimization of energy consumption within hospitals becomes of crucial importance.

Load forecasting plays a vital role in managing buildings and facilities, offering insights with lead-times ranging from minutes to days and months (Abdel-Aal & Radwan, 2004). This aspect of energy management has been shown to have a substantial impact on energy optimization in large buildings (Dagdougui, et al., 2019) including hospitals (Kyriakarakos & Dounis, 2020). Further, accurate predictions of power exchange between on-site renewables and the main power grid are crucial for Smart Grid Management, enabling effective network balance control. This becomes especially significant in micro-grid setups, which are anticipated to be prominent in future power systems. For example, a 2016 studied showed that in aggregated energy systems where intermittent renewables play a significant role, a mere 5% mean absolute percent error in energy demand forecasting can result in substantial unmet demand.

Significant progress has been made in the last decade towards predicting load demand in buildings, driving extensive research efforts in this area. As a result, a variety of models have been proposed to cater to real-world applications (Amasyali & El-Gohary, 2018). In accordance with ASHRAE's classification of energy forecasting models in buildings, there are two categories: physics-based models and data-driven models (Olu-Ajayi, et al., 2022). Physical models rely on the principles of physics and engineering to simulate and predict load demand (Cao, et al., 2020). These models consider factors such as building design, construction materials, thermal characteristics, and weather conditions to estimate energy consumption patterns (Wang, et al., 2019). Several widely used building energy modeling tools have made significant contributions in this area. Notable examples include EnergyPlus, Transient System Simulation Tool (TRNSYS), Environmental Systems Performance-Research (ESP-r), and the Quick Energy Simulation Tool (eQuest) (Bui, et al., 2020). In the realm of physical models, the prediction of building energy consumption involves the consideration of numerous input variables, including a broad range of building-related information, such as the HVAC system specifications, insulation thickness, thermal characteristics, interior occupancy loads, solar data, and other pertinent factors (Cao, et al., 2020). These intensive requirements make physical models better suited for buildings in the design stage rather than for as-built structures (Shao, et al., 2020).

In contrast, data-driven models utilizing artificial intelligence methods to build correlational relationships between given inputs and anticipated outputs have become increasingly prevalent (Wang & Srinivasan, 2017). This rising prominence is largely due to their straightforward application, inherent flexibility, and superior prediction accuracy (Wang, et al., 2019). These models are valued for their ability to handle complex data, demonstrating adaptability across a wide range of industries and applications (Cao, et al., 2020). Furthermore, data-driven models tend to be more applicable than physical based models due to the readily accessible data gathered from buildings, including factors like energy usage, weather conditions, time parameters, occupancy rate, etc., all of which can be efficiently collected through the use of sensors and communication technologies (Wang, et al., 2019).

In hospital facilities, where uninterrupted energy supply is essential, these aspects gain amplified importance, particularly due to continuous usage of technological loads (Bagnasco, et al., 2015). Although the significance of energy demand forecasting is widely acknowledged, its utilization within hospital buildings has received limited attention, particularly in the realm of emerging artificial intelligence (AI) techniques. Recently, AI has demonstrated substantial promise in enhancing energy prediction in buildings by profiling energy consumption, estimating demand patterns, and analyzing sensor data (Dagdougui, et al., 2019). However, achieving these goals requires the development of new machine learning and deep learning approaches. Although the potential of these technologies to mitigate energy consumption and GHG emissions on a worldwide level is significant, there is a notable gap in the existing literature regarding the use of AI and associated data science applications for energy prediction in hospitals setting. This article aims to present a comprehensive literature review focusing on the utilization of AI methodologies in hospital buildings. Optimizing how hospitals use energy is crucial for both cost-saving and environmental reasons. As the world focuses more on sustainable practices, hospitals need to move from traditional methods to modern AI-based techniques for better energy management. There has been significant progress in AI and related technologies to support building operations, but their role in hospitals is still unclear; this research aims to bridge that gap. In this systematic review, we look closely at how data features affect AI's predictions about energy use. By examining the implementation of different AI methods, we hope to clarify how AI can help hospitals predict energy use, guiding research to develop the solutions necessary to improve energy efficiency and decrease carbon emissions for hospitals.

## 2    Literature Review Methodology

This study aims to conduct a comprehensive literature review of articles that employ machine learning and artificial intelligence methods for predicting energy consumption in hospital buildings and is designed to contribute to an in-depth understanding of the current landscape of AI methods in hospital energy prediction. Key elements include the role of various inputs such as occupancy, outdoor temperature, and humidity on energy prediction models, a comparative analysis of various AI techniques used, the intricacies of data preprocessing methods, and the implications of model parameters and hyperparameters.

The methodology of this study is rooted in the PRISMA framework (Figure 1), which involves four stages: identification, screening, eligibility, and inclusion (Sarkis-Onofre, et al., 2021). To guarantee a thorough exploration of the subject, multiple main search phrases along with their related alternatives were employed. The keyword query submitted to diverse databases' search engines was as follows: {(energy optimization*) OR (energy efficiency) OR (energy usage) OR (electricity consumption)} AND {(machine learning) OR (AI) OR (data science) OR (deep learning) OR (data extraction)} AND {(hospital*) OR (medical facilities) OR (healthcare

systems)}. Further, in order to be considered they had to have been written in English, undergone peer review, and been published between 2014 and 2023.

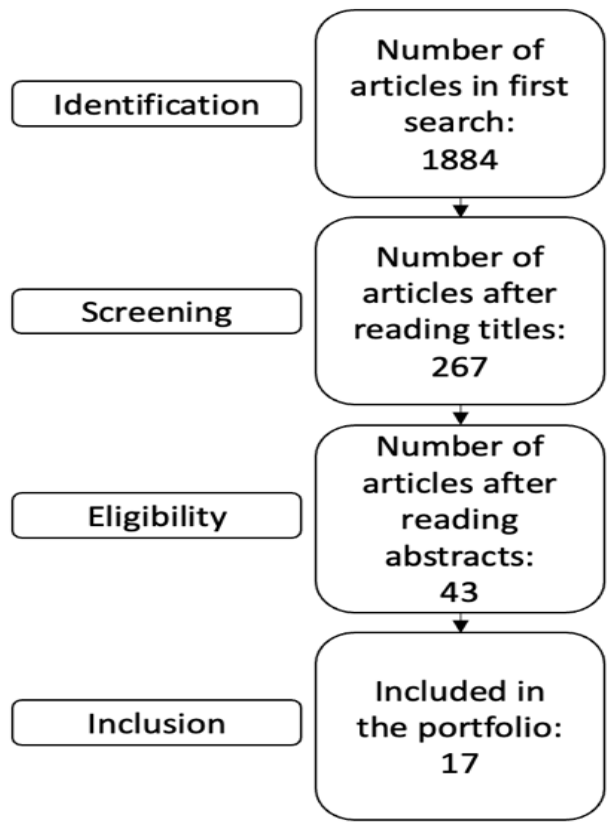

Figure 1: Systematic Literature Review Methodology using PRISMA

Based on these criteria, an initial 1884 publications were considered. In the title screening phase, the primary focus was to ascertain whether the title indicated the application of AI and machine learning in the context of energy optimization. Furthermore, any publication with a title suggesting a focus on topics other than hospital buildings was deliberately excluded from further consideration. After this screening process, 267 articles remained. In the next stage, the abstracts of each of these selected publications were subjected to a detailed review, leading to the identification of 43 articles suitable for a comprehensive full-paper assessment. During this final stage of review, 26 articles were found to be unrelated to the research topic and were consequently excluded. This paper aims to summarize the findings and insights gleaned from the remaining 17 selected articles, which form the basis of our study.

In order to structure our literature review, the selected papers were analyzed based on the following research questions:

1. How does the selection of features, such as occupancy data, meteorological data, and time dynamics, influence the performance and accuracy of AI models in energy forecasting?
2. What are the crucial data preprocessing and feature engineering steps that optimize machine learning algorithms for energy forecasting?
3. Which AI techniques are prominently used in energy prediction, and how do they vary in efficacy and applicability based on the underlying data and features?

The next section will delve into these literature findings, providing a cohesive and structured presentation of the knowledge obtained from the selected articles. By synthesizing this information, we aim to contribute to an in-depth understanding of the interaction between machine learning, artificial intelligence, and energy optimization in hospital settings. With this comprehensive approach, we hope to pave the way for future advancements in AI-driven energy prediction, furthering our collective efforts toward a greener and more energy-efficient healthcare infrastructure.

## 3  Literature Review Findings

The findings of the literature review are presented as-follows. First, the varying timescales for energy forecasting and their unique requirements and common approaches are discussed. Next, the most common features are discussed and their impact when included in models is presented. Finally, the range of machine learning approaches are summarized with respect to data preprocessing, feature engineering, and the range of algorithms and artificial intelligence techniques.

### 3.1 Time Scale Variations in Energy Forecasting

Energy forecasting serves different purposes based on the time horizon considered. Short-term forecasting focuses on predicting energy demand in the upcoming minutes, hours, or days. Its significance lies in enabling power generation companies to align production operations with market demands, allowing distribution companies to optimize grid adjustments accordingly, and helping buyers strategize their purchasing decisions for optimal prices (Hong & Fan, 2016). Indeed, Short-Term Energy Forecasting (STEF) plays a critical role in various aspects of the energy industry. Firstly, it facilitates the optimal unit commitment, enabling power generation companies to make informed decisions about which units to commit for production based on the forecasted energy demand. This helps ensure efficient utilization of resources and minimizes costs (Raza & Khosravi, 2015). Furthermore, STEF aids in the control of spinning reserves, which are additional power generation capacities held in reserve to quickly respond to sudden increases in demand or unexpected outages. Accurate short-term forecasts allow for effective management of spinning reserves, ensuring they are available when needed while minimizing unnecessary usage and associated costs (Fernández-Martínez & Jaramillo-Morán, 2022). Moreover, STEF contributes to the evaluation of sales or purchase contracts between different companies in the energy market. By providing accurate and timely predictions of energy demand, companies can negotiate contracts based on reliable forecasts, optimizing their sales and purchases to meet the anticipated market needs efficiently (Timur, et al., 2020).

For medium-term predictions, the goal shifts to estimating energy demand over the following weeks or months. Medium-Term Electrical Energy Forecasting (MTEF) plays a crucial role in optimizing the operation and maintenance of large-scale power system facilities (Bekteshi, et al., 2015). By accurately predicting energy demand over the medium-term, MTEF enables companies and state facilities to align production schedules, capacity planning, and resource utilization with market needs. This optimization prevents overproduction or underproduction scenarios, maximizing efficiency and profitability (Fernández-Martínez & Jaramillo-Morán, 2022). Additionally, MTEF assists in planning maintenance schedules, minimizing disruptions, and improving operational efficiency. It also supports effective management of fuel supplies, ensuring continuous and reliable availability to avoid operational disruptions (Fernández-Martínez & Jaramillo-Morán, 2022). By leveraging MTEF, organizations can strategically plan investments in renewable energy technologies, evaluating their feasibility and economic viability for integration into the energy portfolio. Ultimately, MTEF empowers entities to make informed decisions, optimize their energy operations, and work towards sustainability goals (Fernández-Martínez & Jaramillo-Morán, 2022).

Lastly, long-term prediction concentrates on forecasting the overall annual energy demand and load peaks for future years (Timur, et al., 2020). Long Term Energy Forecasting (LTEF) plays a pivotal role in long-term power system planning, aligning with a country's future energy demand and policy. It provides valuable insights into anticipated energy consumption trends over an extended period, enabling policymakers, energy authorities, and utility companies to make informed decisions (Fernández-Martínez & Jaramillo-Morán, 2022). By accurately forecasting long-term energy demand, LTEF assists in identifying potential gaps in energy supply and determining the necessary infrastructure investments needed to meet future demand. This information helps shape comprehensive power system plans that align with the evolving energy landscape (Timur, et al., 2020). Moreover, LTEF informs the development of energy policies at a national level, enabling policymakers to formulate robust strategies that ensure reliable and sustainable energy supply. These policies may include promoting renewable energy integration, implementing energy efficiency initiatives, and encouraging the adoption of clean technologies (Raza & Khosravi, 2015). These categories enable a comprehensive understanding and classification of electrical energy forecasting methods based on their respective time horizons. Table 1 classifies thee reviewed studies by time scale for energy prediction. In this table, there is a significant focus on short-term and very short-term forecasting. This shift towards shorter forecast horizons is potentially driven by the increasing need for dynamic energy management and optimization in rapidly evolving energy markets.

Table 1 Time Scales Used in the Reviewed Papers

| Forecast Type | Time Scale | References |
|---|---|---|
| **Medium-Term Energy Forecasting (n = 2)** | One month ahead | (Timur, et al., 2020) |
| | One week ahead | (Cao, et al., 2020) |
| **Short-Term Energy Forecasting (n = 13)** | Two days ahead | (Ngo, et al., 2021) |
| | One day ahead | (Ruiz, et al., 2017), (Bagnasco, et al., 2015), (Cao, et al., 2020) |
| | 30 hours ahead (STEF) | (Nakai, et al., 2021), (Ozaki, et al., 2021) |
| | 24 hours ahead | (Fernández-Martínez & Jaramillo-Morán, 2022), (Manno, et al., 2022), (Ngo, et al., 2021) |
| **Very Short-Term Energy Forecasting (n = 7)** | 1 hour ahead | (Zini & Carcasci, 2023), (Zor, et al., 2020), (Panagiotou & Dounis, 2022), (Buluş & Zor, 2021) |
| | Real time | (Zini & Carcasci, 2023), (Maddalena, et al., 2022), (Yang & Wan, 2022) |

## 3.2 Impact of Feature Selection on Model Performance

The success of AI models in forecasting time series data depends on the quality and the appropriateness of selected features in the input data (Timur, et al., 2020). Considering the significance of energy efficiency in buildings, it becomes essential to examine the factors that shape energy usage, as highlighted in a project report by the International Energy Agency (IEA). These factors, encompassing climate, building structure, equipment, operational practices, occupant behavior, and indoor environmental conditions, play a vital role in determining energy consumption (Cao, et al., 2020). Healthcare buildings, in particular, present a unique set of challenges in terms of energy consumption when compared to other commercial facilities. Due to their hybrid nature, serving as both general-purpose spaces and specialized medical units, additional factors come into play, further influencing their energy usage patterns (Fernández-Martínez & Jaramillo-Morán, 2022). An examination of the literature reveals a diversity of data inputs leveraged for optimizing energy in hospital buildings. These data inputs commonly fall into four categories: meteorological, date and time, occupancy, and operational data. Their application across various studies is comprehensively compiled in Table 2.

Table 2 Data Types Used in the Literature

| Data Type | References |
|---|---|
| **Meteorological Data** | |
| **Outdoor temperature (Mean, Max, Minimum) (n=16)** | (Timur, et al., 2020), (Ruiz, et al., 2017), (Cao, et al., 2020), (Ozaki, et al., 2021), (Nakai, et al., 2021), (Zor, et al., 2020), (Fernández-Martínez & Jaramillo-Morán, 2022), (Manno, et al., 2022), (Bagnasco, et al., 2015), (Zini & Carcasci, 2023), (Xue, et al., 2020), (Buluş & Zor, 2021), (Yang & Wan, 2022), (Chalapathy, et al., 2021), (Maddalena, et al., 2022), (Ngo, et al., 2021) |
| **Relative Humidity (n=6)** | (Timur, et al., 2020), (Cao, et al., 2020), (Zor, et al., 2020), (Bagnasco, et al., 2015), (Buluş & Zor, 2021), (Fernández-Martínez & Jaramillo-Morán, 2022) |
| **Wind Speed (n=4)** | (Timur, et al., 2020), (Cao, et al., 2020), (Zor, et al., 2020), (Buluş & Zor, 2021) |
| **Solar radiation (n=3)** | (Zor, et al., 2020), (Manno, et al., 2022), (Xue, et al., 2020) |
| **Precipitation (n=3)** | (Zor, et al., 2020), (Buluş & Zor, 2021), (Cao, et al., 2020) |
| **Barometric pressure (n=2)** | (Cao, et al., 2020), (Buluş & Zor, 2021) |
| **Wind direction (n=2)** | (Zor, et al., 2020), (Buluş & Zor, 2021) |
| **Precipitation (n=1)** | (Cao, et al., 2020) |
| **Sunshine duration (n=1)** | (Timur, et al., 2020) |
| **Date and Time** | |
| **Date (n=8)** | (Manno, et al., 2022), (Buluş & Zor, 2021), (Ruiz, et al., 2017), (Zor, et al., 2020), (Nakai, et al., 2021), (Ozaki, et al., 2021), (Bagnasco, et al., 2015), (Zini & Carcasci, 2023) |
| **Day of Week/ Holiday value (n=6)** | (Zor, et al., 2020), (Manno, et al., 2022), (Cao, et al., 2020), (Ozaki, et al., 2021) (Buluş & Zor, 2021), (Bagnasco, et al., 2015), |
| **Time (n=5)** | (Nakai, et al., 2021), (Bagnasco, et al., 2015), (Zini & Carcasci, 2023), (Buluş & Zor, 2021), (Zor, et al., 2020) |
| **Occupancy** | |
| **Number of Patients** (n=1) | (Cao, et al., 2020) |
| **Number of Outpatients** (n=1) | (Cao, et al., 2020) |
| **Number of Emergency Patients** (n=1) | (Cao, et al., 2020) |

| **Operational data** | |
|---|---|
| **Historical electricity consumption** (n=12) | (Timur, et al., 2020), (Cao, et al., 2020), (Ozaki, et al., 2021)1, (Nakai, et al., 2021), (Zor, et al., 2020), (Ruiz, et al., 2017), (Buluş & Zor, 2021), (Fernández-Martínez & Jaramillo-Morán, 2022), (Manno, et al., 2022) (Xue, et al., 2020), (Bagnasco, et al., 2015), (Zini & Carcasci, 2023) |
| **Historic chiller operational data** (n=1) | (Chalapathy, et al., 2021) |
| **Hourly load** (n=1) | (Panagiotou & Dounis, 2022) |
| **Operational status of central air conditioning system** (n=1) | (Cao, et al., 2020) |
| **Cooling delivery system flows** (n=1) | (Chalapathy, et al., 2021) |

As observed from the above table, the most widely-used data types (i.e. input variables) are outdoor air temperature, historical electricity consumption, and time and date values. This prevalence can be attributed to the direct impact these variables have on energy consumption patterns: outdoor air temperature influences the demand for heating and cooling systems, historical electricity consumption provides a baseline that assists in predicting future usage, and time and date values allow models to account for periodic variations and trends in energy use, which are pivotal in forming accurate forecasts in various AI models. In order to develop precise and efficient AI and machine learning models for energy optimization in healthcare buildings, a comprehensive understanding of these input variables is essential. There are a number of factors that influence the energy load in healthcare facilities and each of these factors affects it in a different manner. Many studies included in the review did not offer an exhaustive analysis or an in-depth discussion of their input data. However, there were a few studies that examined the significance of each variable employed in their models (Cao, et al., 2020; Timur, et al., 2020; Ruiz, et al., 2017). To better understand the most common input variables used in the literature for studying energy consumption in healthcare buildings, they are discussed in the following sections. By delving into each of these features in greater detail within each category (occupancy data, meteorological data, operational data, date and time data), we aim to illustrate their relevance and impact on energy consumption.

### 3.2.1 Occupancy Data

Occupancy is one of the key factors that influences energy consumption in healthcare facilities (Cao, et al., 2020). Various studies have illustrated the significant influence occupancy has on energy demand patterns. The majority of research in healthcare buildings on energy prediction pays more attention to days of the week or the type of day (weekday or weekend/holiday) as a surrogate measure for occupancy level in short term energy consumption (Ruiz, et al., 2017; Buluş & Zor, 2021; Ngo, et al., 2021; Zor, et al., 2020; Manno, et al., 2022; Cao, et al., 2020; Ozaki, et al., 2021; Buluş & Zor, 2021; Bagnasco, et al., 2015). These studies found that hospitals' activities, and thus energy consumption, decrease significantly during weekends and holidays due to reduced occupancy and services. However, it's worth noting that hospitals operate around the clock, and thus the day type might have less influence in the case of healthcare buildings as compared to other commercial buildings (Cao, et al., 2020). To encapsulate the complexity of occupancy, Ruiz et al. (2017) included additional features representing daily hospital activities like admissions, discharges, revisions, and emergencies among others in their daily load prediction models. They found that using these activity-based features did improve prediction accuracy, but not as substantially as expected ($R^2 = 0.943$). This moderate improvement could be attributed to two primary reasons. Firstly, the performance of models using only the day type was already decent ($R^2 = 0.914$), as it indirectly encapsulates information about hospital activities, most of which run only on workdays, through the weekday/weekend variable. Secondly, the energy consumption data is generally collected at a building level, but activities like emergencies, medical tests, and surgical interventions happen at specific locations within the building. Hence, the global energy consumption data may not fully reflect the impact of these localized activities (Ruiz, et al., 2017). The findings from these studies suggest that in the absence of detailed daily activity information, which can be challenging and costly to collect, the type of day (weekday/weekend/holiday) can be a reasonably effective predictor of hospital activity and consequently energy consumption. A counterintuitive finding of one study (Timur, et al., 2020) was that the number of patients remained steady over the years and did not significantly influence medium-term electrical energy forecasting. This finding may indicate that, in the context of this specific study, hospital occupancy didn't significantly impact medium or long-term energy consumption forecasts. However, it's important to consider that this may not hold true in scenarios where patient numbers fluctuate significantly. Changes in hospital usage, driven by political or business decisions, urban growth, or other factors, could each potentially alter the energy consumption patterns over the medium to long term.

The people occupying a hospital are not homogeneous and includes different groups like doctors, nurses, patients, visitors, and facilities managers, each undertaking different levels of activity or with different clothing levels that could impact both building internal loads and occupant comfort. In a healthcare setting, different departments exhibit distinctive energy usage patterns which could be attributed to their varied functions and occupancy profiles (Cao, et al., 2020). For example, outpatient departments, which mainly focus on diagnosing and treating patients who do not require hospitalization, primarily consume energy through HVAC systems, lighting, and medical equipment. In contrast, inpatient departments, which cater to hospitalized patients, additionally consume energy for domestic hot water systems, catering, and cooking appliances (Cao, et al., 2020). Most previous studies have failed to consider these department-specific occupancy types.

Only one study (Cao, et al., 2020) included the number of in-patients and out-patients as inputs to their model, but did not discuss the impact of this inclusion. Further research is needed to analyze the effects of occupancy by expanding occupancy considerations to consider different occupant types.

In conclusion, occupancy is a complex and multifaceted factor that influences energy consumption in healthcare facilities. While some dimensions of occupancy have been explored, there is a need for more detailed and nuanced investigations, especially around the role of department-specific occupancy types, the influence of different occupant groups, and the impact of daily hospital activities on energy consumption. In the absence of such detailed information, day-type can serve as a reasonable proxy for occupancy level. However, future research should aim to uncover more granular and comprehensive insights about occupancy and its relationship with energy consumption in healthcare settings.

### 3.2.2 Meteorological Data

A vast array of research literature underscores the significance of weather parameters in influencing energy consumption in complex buildings such as hospitals (Buluş & Zor, 2021; Cao, et al., 2020; Zini & Carcasci, 2023; Zor, et al., 2020; Timur, et al., 2020; Ruiz, et al., 2017; Bagnasco, et al., 2015; Manno, et al., 2022). The weather parameters commonly include outdoor temperature, relative humidity, wind speed, barometric pressure, and precipitation (Wang, et al., 2019). These variables serve as the primary drivers of energy demand, particularly due to their influence on the operation of Heating, Ventilation, and Air Conditioning (HVAC) systems, which account for sixty percent of energy use in such establishments (Cao, et al., 2020).

Among these variables, outdoor temperature has shown a strong correlation with energy consumption in healthcare facilities (Ruiz, et al., 2017; Cao, et al., 2020; Bagnasco, et al., 2015;

Zor, et al., 2020). This correlation is evident from its consistent inclusion as an independent variable in all reviewed studies, except for one study that only utilized the total energy load of previous hours as input (Panagiotou & Dounis, 2022). This focus on outdoor temperature is further highlighted in another study which elucidated the differential impact of external temperature on summer and winter energy consumption (Zini & Carcasci, 2023). The study found that while summer electricity consumption was markedly affected by external temperature, the winter energy demand was less influenced, as their hospital was served primarily by thermal (central hot water and steam), rather than electrical sources. Their findings also emphasized the dependency of energy demand on the seasonal changes of the outdoor climate, indicating the importance of a comprehensive understanding of climate parameters in energy demand characterization.

Interestingly, the same study (Zini & Carcasci, 2023) pointed to the presence of multicollinearity among the climate parameters, especially a strong negative correlation between outdoor temperature and relative humidity. Although this correlation suggests relative humidity decreases as temperature rises, the study cautioned that the daily behaviour of these two variables might present a different correlation, thus necessitating their consideration as input features in machine learning models. A study conducted in Shanghai (Cao, et al., 2020) explored this complexity further, finding a U-shaped relationship between outdoor temperature and daily electricity consumption. Their investigation further corroborated the importance of outdoor temperature, with it being identified as the most influential factor on energy consumption in hospital buildings, followed by pressure.

Contrary to the primary focus on temperature, another study (Ruiz, et al., 2017) posited that predicting daily electrical energy consumption based on temperature variables alone was not sufficient, thereby suggesting the need for inclusion of other weather variables like humidity and radiation. While the correlation between humidity and energy load was found to be low in one study (Bagnasco, et al., 2015), thus leading to its exclusion in the final dataset, other research (Manno, et al., 2022) indicated a moderate correlation between humidity and energy demand, indicating the importance of considering this variable in the research landscape. Beyond temperature and humidity, short-wave irradiation and wind speed have also been identified as influential factors in energy consumption (Zor, et al., 2020; Manno, et al., 2022) . In particular, the incorporation of short-wave irradiation in the prediction model was found to significantly increase the complexity of short-term building electrical energy consumption prediction (Zor, et al., 2020).

In conclusion, while outdoor temperature is indeed a major influencer on energy consumption in complex structures like hospitals, the multifaceted interplay of weather parameters such as humidity, short-wave radiation, and wind speed also demands careful attention. The relationships between these variables and energy demand are complex, often non-linear (Zini & Carcasci, 2023), and their combined effect can vary widely depending on the specifics of the building and the surrounding environment (Cao, et al., 2020). While the current body of research provides valuable insights into these relationships, there are several avenues for future exploration to further refine our understanding and prediction of energy consumption. For instance, there is a need for more comprehensive studies that systematically investigate the relative influence of a broader range of weather parameters on energy consumption beyond just temperature, taking into account the potential for complex interactions and non-linear relationships between these variables. Furthermore, while the existing research identifies some key variables such as short-wave radiation (Zor, et al., 2020) and humidity (Manno, et al., 2022) ,the impact of others, like wind speed and barometric pressure, on energy consumption has not been fully explored. Future research could therefore aim to understand how these lesser-studied weather parameters might influence energy consumption, either directly or through their interaction with other variables. Additionally, more research could be done to understand the specific daily behaviors of key variables like temperature and humidity (Zini & Carcasci, 2023) and their influence on energy consumption. Currently, there is a gap in understanding of how these variables interact on a daily scale and how this might impact energy consumption predictions. The continued exploration of these research directions will enhance our understanding of the intricate relationships between weather data and energy consumption, ultimately leading to more accurate and reliable prediction models for energy management in complex structures like hospitals.

### 3.2.3 Daily and Seasonal Variations

In addition to the day of the week, another important aspect of energy consumption analysis lies in understanding the time dynamics and operational status of the healthcare facilities. The importance of time dynamics, particularly daily and seasonal variations, has been highlighted in a number of studies (Zini & Carcasci, 2023; Zor, et al., 2020; Timur, et al., 2020; Cao, et al., 2020; Manno, et al., 2022; Fernández-Martínez & Jaramillo-Morán, 2022). Meanwhile, the operational status of a facility is an aspect that has been less thoroughly explored, but its relevance has been identified in one study (Cao, et al., 2020).

Time dynamics refers to variations in energy consumption patterns over different periods, from intra-day to inter-seasonal scales (Zor, et al., 2020). Energy consumption in hospitals does not remain constant throughout the day. For example, Zini and Carcasci (2023) observed distinctive daily average power curves for each season, indicating an interplay of seasonal factors and daily operational patterns. This study highlights how peak energy consumption corresponds with periods of high healthcare activity intensity, with a progressive growth of demand occurring between 06:00 a.m. to 08:30 a.m. and the most energy-intensive period taking place from 08:30 a.m. until 6:00

p.m. However, these trends are influenced by the season, with summer, for instance, showing a nearly linear decrease in electricity demand after peak hours, presumably due to reduced operation of air handling systems. The significant influence of time on energy consumption was also evident on a larger scale. Several studies (Zor, et al., 2020; Timur, et al., 2020; Fernández-Martínez & Jaramillo-Morán, 2022) reported clear seasonality effects, with peak consumption observed during extreme weather months. In particular, Fernández and Jaramillo (2022) showed consumption peaks during January and February, and, more significantly, during the summer months. This seasonal effect is closely tied to weather patterns, indicating an interrelation between weather and time dynamics, as previously discussed in the weather data section.

In terms of the operational status of healthcare facilities, it's highly important to consider the operation and maintenance (O&M) measures related to the central air-conditioning system. According to one study (Cao, et al., 2020), the operational status of the central air conditioning system, including whether it is on or off, and if on, whether it is heating or cooling, significantly impacts electricity consumption. The decision to turn on or off the central air conditioning system is often based on a predefined outdoor temperature threshold, thus introducing a non-linear relationship with temperature. In other words, the energy consumption does not increase or decrease linearly with temperature; rather, it depends on the O&M policy. Interestingly, this study showed that the operational status of the air conditioning system ranked third in terms of importance, after outdoor temperature and pressure.

In light of these findings, it's apparent that time dynamics and operational status play significant roles in energy consumption analysis. Despite the insights gained from these studies, gaps remain that future research should address. Firstly, only one study (Cao, et al., 2020) has extensively discussed the operational status and its impact on energy consumption, suggesting a need for more research in this area. A more in-depth understanding of how various O&M measures affect energy consumption could offer valuable insights for energy management strategies. Additionally, more research is needed to understand the interactions between weather parameters, time dynamics, and operational status, as these aspects are likely to influence each other. Furthermore, more comprehensive models that capture daily and seasonal fluctuations, as well as operational status, should be developed to enhance the accuracy of energy consumption prediction. The studies (Zor, et al., 2020; Timur, et al., 2020) show that model accuracy tends to decline at the start and end of shifts and during seasonal changes, indicating these are areas that could benefit from more detailed modeling.

Overall, understanding the interplay of time dynamics, operational status is essential for building accurate and reliable energy consumption prediction models. Future research needs to bridge the existing gaps by focusing on these less-explored areas, which will ultimately aid in the development of more efficient energy management strategies for complex structures such as hospital.

### 3.3  Machine Learning Algorithms and Their Implementation

In order to comprehensively explore the literature regarding the role, implementation, and impact of various machine learning algorithms in the context of energy forecasting, data pre-processing must be considered alongside the algorithms themselves. Section 3.3.1 underscores the importance of data preprocessing and feature engineering, highlighting the influence of these processes on the accuracy and reliability of forecasting models. Following this, Section 3.3.2, presents a detailed examination of various machine learning algorithms used in energy prediction.

### 3.3.1 Data Preprocessing and Feature Engineering in Energy Forecasting

The role of data preprocessing and feature engineering in energy forecasting is instrumental, shaping the accuracy and reliability of forecasting models by refining and structuring raw data into meaningful insights (Nakai, et al., 2021). However, despite the critical nature of these processes, a significant portion of the literature doesn't detail their preprocessing strategies. This gap underscores a broader challenge in the domain, emphasizing the need for transparent and exhaustive reporting for replication and broader understanding.

One of the primary aspects to confront during preprocessing is addressing missing values. From the reviewed literature, multiple strategies emerge, underlining the importance of context. From the reviewed literature, multiple strategies emerge, underlining the importance of context. Several studies (Ozaki, et al., 2021; Chalapathy, et al., 2021; Fernández-Martínez & Jaramillo-Morán, 2022) explore the diversity of these methods. While Ozaki et al. (2021) opted to fill gaps with preceding measurements, Fernández-Martínez and Jaramillo-Morán, (2022) utilized interpolation based on the dataset's periodic behavior and Chalapathy et al. (2021) discarded data points that indicate sensor malfunctions. Each approach was tailored to the specific nature of their dataset, emphasizing that context greatly influences the strategy for managing inconsistencies. This processing doesn't stop at handling irregularities. Many machine learning models, especially neural networks and SVRs, display sensitivity to data scales, making normalization a pivotal step (Ngo, et al., 2021). The Min-Max normalization method, which transforms numerical data values into a specific range, often between zero and one, using the minimum and maximum values in the dataset, was adopted in multiple studies (Fernández-Martínez & Jaramillo-Morán, 2022; Timur, et al., 2020; Ngo, et al., 2021) and stands out as a commonly employed technique. But while its recurring use indicates a trusted method, it raises a question: *could alternative normalization methods provide varied, potentially more accurate, results?* This hints at a potential avenue for further exploration, comparing the impacts of different normalization techniques on prediction accuracy.

The process of refining data continues with feature engineering, aimed at improving the model's ability to comprehend and predict intricate patterns (Ozaki, et al., 2021). Feature engineering may be conducted prior to, concurrently with, or after the feature selection process. However, it is typically undertaken as the final step in the analysis. This is mainly because the number of features can be substantial, and conducting a systematic analysis involves considerable computational time that might not be feasible in a practical setting (Zini & Carcasci, 2023). Illustrating this, studies like (Zini & Carcasci, 2023) and (Panagiotou & Dounis, 2022) transformed hourly patterns using sine and cosine functions, which elegantly captures cyclic behaviors inherent to energy consumption. Additionally, the 'moving-window approach' , a methodology adopted and validated in time-series forecasting (Ngo, et al., 2021), stresses the importance of formatting time-series data suitably for supervised learning, reiterating that while some methods may be traditional, their efficiency remains undiminished.

Integrated within these processes is the powerful tool of correlation analysis. Leveraging the Pearson correlation coefficient, as seen in (Manno, et al., 2022), enables researchers to pinpoint influential features, in this case, specific weather data components that drastically affect heating or electricity demand. However, it's vital to recognize the limitations of a linear correlation measure in a world teeming with non-linear relationships. Hence, the introduction of advanced feature selection mechanisms, such as the GMDH polynomial neural networks highlighted by

Buluş and Zor (2021), signifies the merging of deep learning techniques in feature engineering. It is important to note that deep learning is not only capable of reducing dimensionality, but it can also enhance a model's efficacy by focusing on crucial predictors (Zor, et al., 2020).

Further deepening the layers of preprocessing, one study (Fernández-Martínez & Jaramillo-Morán, 2022) introduced the concept of time-series decomposition. By breaking down time-series data into manageable subseries, forecasting might become more straightforward. Still, one can't help but question the universal applicability of this method. Would such a decomposition consistently augment model accuracy, or might it inadvertently complicate the process and pave the way for overfitting? Reviewing the literature reveals a notable gap: many studies don't provide detailed methodologies for preprocessing. This gap in knowledge makes it challenging to replicate studies and compare findings across different research. Nonetheless, a consistent theme emerges: the quality of data is directly linked to the accuracy of the model. While the reviewed articles provide valuable insights into preprocessing and feature engineering for energy forecasting, there's a clear need for more transparency and structured comparisons in the field. As research continues, the emphasis on proper data preparation and its relationship to accurate forecasting is undeniably pivotal.

### 3.3.2 Overview of AI Techniques in Energy Prediction

Machine learning has become increasingly prevalent in various fields, including energy consumption forecasting (Zor, et al., 2020). In this content, numerous machine learning methods have been explored to optimize forecasting accuracy. Machine learning, fundamentally, is the computational process of enabling machines to learn from data, thereby facilitating them to discern patterns and make informed future predictions (Nakai, et al., 2021). Various studies have employed algorithms from traditional regression models like Multiple Linear Regression (MLR) (Zini & Carcasci, 2023) to more sophisticated ones such as deep neural networks (Nakai, et al., 2021).

Delving into the comparison of different methods, several studies have highlighted the strengths and limitations of various algorithms. Table 3 summarizes different algorithms used in the literature.

Table 1: AI models used in the literature

| Algorithms | References |
|---|---|
| **Artificial Neural Networks (ANNs)** (n=9) | (Timur, et al., 2020), (Ozaki, et al., 2021), (Zor, et al., 2020), (Ruiz, et al., 2017), (Chalapathy, et al., 2021), (Panagiotou & Dounis, 2022), (Manno, et al., 2022), (Bagnasco, et al., 2015), (Zini & Carcasci, 2023) |
| **Support Vector Regression (SVR)** (n=5) | (Cao, et al., 2020), (Chalapathy, et al., 2021), (Ngo, et al., 2021), (Manno, et al., 2022), (Xue, et al., 2020) |
| **Deep Neural Network (DNN)** (n=3) | (Nakai, et al., 2021), (Buluş & Zor, 2021), (Yang & Wan, 2022) |
| **Long Short-Term Memory (LSTM)** (n=3) | (Nakai, et al., 2021), (Panagiotou & Dounis, 2022), (Fernández-Martínez & Jaramillo-Morán, 2022) |
| **Random Forest (RF)** (n=3) | (Nakai, et al., 2021), (Cao, et al., 2020), (Xue, et al., 2020) |
| **Auto-Regressive Integrated Moving Average (ARIMA)** (n=2) | (Panagiotou & Dounis, 2022), (Manno, et al., 2022) |
| **Gaussian Process** (n=2) | (Cao, et al., 2020), (Maddalena, et al., 2022) |
| **Linear Regression** (n=2) | (Timur, et al., 2020), (Cao, et al., 2020) |
| **XGBoost Random** (n=2) | (Cao, et al., 2020), (Chalapathy, et al., 2021) |
| **Decision Tree** (n=1) | (Xue, et al., 2020) |
| **Gated Recurrent Unit (GRU)** (n=1) | (Fernández-Martínez & Jaramillo-Morán, 2022) |
| **Lasso Regression, Gaussian Process, Ridge Regression** (n=1) | (Cao, et al., 2020) |
| **Naïve Linear Extrapolation** (n=1) | (Chalapathy, et al., 2021) |
| **Support Vector Machine (SVM)** (n=1) | (Timur, et al., 2020) |

Historically, methods such as Multiple Linear Regression (MLR) were prevalent in these endeavors (Zini & Carcasci, 2023). Although MLR has been influential, it exhibits constraints, particularly when confronted with complex datasets. The limitations of MLR paved the way for more advanced tools like the Artificial Neural Network (ANNs) (Zini & Carcasci, 2023). In comparing these two methods, while MLR yielded valuable insights, ANN emerged as the superior tool for forecasting, particularly in distinguishing distinctive energy usage patterns during morning and afternoon periods (Zini & Carcasci, 2023). However, given the multifaceted nature of energy prediction, a singular model or methodology might be inadequate. This led to the inception of ensemble methods where models like Extreme Gradient Boosting (XGBoost) and Random Forest (RF) are integrated for enhanced accuracy (Cao, et al., 2020). The collective intelligence of ensemble methods often results in more comprehensive predictions.

The efficiency of these models is not solely dependent on their architecture but also on their parameters. Refining parameters such as the learning rate, dropout rate, or the number of hidden layers holds the potential to greatly enhance model performance (Ozaki, et al., 2021; Ngo, et al., 2021; Nakai, et al., 2021). Ozaki et al. (2021) tested three tuning methods—Grid Search, Random Search, and Bayesian Optimization—and found all to enhance prediction accuracy, with Bayesian Optimization showing consistent improvements, although not surpassing Auto ML (Ozaki, et al., 2021). Moreover, in pursuit of enhancing prediction models, researchers have ventured into hybrid models. The integration of Support Vector Regression (SVR) with the Grey Wolf Optimizer (GWO) exemplifies this innovative approach (Ngo, et al., 2021). The SVR serves as the prediction mechanism while the GWO optimizes its hyperparameters. Notably, the Wolf-Inspired Optimized Support Vector Regression model consistently outperformed other traditional machine learning models, presenting a promising avenue for future research in optimization algorithms (Ngo, et al., 2021).

However, it's important to note that while accuracy is a crucial goal, the temporal dimension of decision-making underscores the significance of speed in predictive models (Bagnasco, et al., 2015). The emphasis on quick decision-making in predictive models is highlighted by techniques like Instantaneous Linearization (IL) (Bagnasco, et al., 2015). When incorporated into machine learning (ML)-based model predictive control (MPC) for energy management, IL significantly decreased CPU processing time by approximately 98.6%, reducing it from 15.64 seconds to an average of 0.22 seconds, although MPC's high computational demand remains a challenge (Shiyu & Wan, 2022).

When evaluating various machine learning algorithms for load forecasting, SVM stood out for its high accuracy but was encumbered by extended computational times (Timur, et al., 2020). In contrast, Multilayer Perceptron (MLP) networks displayed a balanced performance with high accuracy combined with rapid run times (Timur, et al., 2020).While MLP offered the swiftest results, its accuracy was suboptimal. If the objective is minimal error, SVM remains unparalleled; however, for a blend of accuracy and speed, MLP is recommended (Timur, et al., 2020). Additionally, a particular focus on ensuring optimal computational speed led to the selection of a single hidden layer for the ANN model (Bagnasco, et al., 2015). Even though the model exhibited potency across simulations, it faced reduced accuracy during peak load times (Bagnasco, et al., 2015). Consequently, addressing these accuracy drops during peak loads necessitates further optimization of the model, as emphasized by the same authors.

Emerging methodologies, such as the Group Method of Data Handling (GMDH) networks, further enrich the field of energy prediction (Zor, et al., 2020). Every novel research endeavor, whether refining established models such as GMDH (Buluş & Zor, 2021) or pioneering new avenues like Recurrent Neural Network Multi-Input Multi-Output(RNN-MIMO) models (Chalapathy, et al., 2021), augments the collective understanding of energy consumption patterns.

The findings of the reviewed articles highlight the crucial role that machine learning and deep learning models play in predicting and managing energy consumption. The selection of a particular model depends on specific requirements, such as accuracy, interpretability, and computational speed. Hybrid models and ensemble approaches often outperform individual models, and the optimization of hyperparameters can significantly enhance prediction accuracy. Yet, as the exploration for the perfect blend of precision and efficiency continues, addressing challenges such as peak load times and integrating emerging methodologies remain paramount. This intricate balance will not only shape the future of energy forecasting but also significantly influence the operational efficiency of critical infrastructures like hospitals.

## 4   Discussion

Healthcare buildings, a subset of commercial structures, exhibit unique energy consumption patterns, primarily due to their dual nature of housing both general-purpose spaces and specialized medical units. Delving into literature, we notice a diversity in the data inputs that researchers utilize. These data types range from meteorological inputs to occupancy, time variations, operational status, and even the raw data upon which predictive models are based. Such diversification suggests that to predict energy consumption accurately, a comprehensive appreciation of these variables is paramount. Interestingly, while many studies elucidate these variables, few delve deep into the intricacies and implications of their choices.

Occupancy, for instance, emerges as a paramount factor in energy usage in hospitals. A substantial volume of research underscores how different days of the week or types of days can serve as a proxy for occupancy level. Hospitals, with their round-the-clock operations, exhibit a different energy consumption pattern during weekends and holidays, primarily driven by decreased activities. Yet, there's complexity to consider. Not all activities in a hospital have the same energy impact. For example, admissions or emergencies might not always reflect on the global energy consumption scale. This calls for an exploration of more granular data, like the types of occupants or the specific daily hospital activities, to refine prediction accuracy.

Similarly, the role of meteorological data in energy consumption forecasting cannot be understated. Outdoor temperature, for example, consistently emerges as a potent predictor of energy usage in hospitals. It's evident from numerous studies that temperature has a pronounced effect on energy consumption, particularly during summer. However, like occupancy, the relationship between meteorological data and energy consumption isn't linear. While temperature remains paramount, there are hints in the literature about the significance of other factors like humidity, short-wave radiation, and even wind speed. This raises the question: How do these variables interplay on a daily scale, and what implications does this have for prediction accuracy?

Moreover, incorporating the aspect of forecasting timescales, it's imperative to recognize how these timescales - short-term, medium-term, and long-term - play a crucial role in energy management within healthcare facilities. Short-term forecasting, for instance, is vital for day-to-day operational efficiency, allowing hospitals to adjust their energy usage in response to immediate demand changes. Medium-term forecasting becomes relevant for maintenance scheduling and resource planning, enabling hospitals to prepare for upcoming seasonal variations or planned events. Long-term forecasting, on the other hand, is crucial for strategic planning and investment in energy infrastructure. By understanding these varying timescales, healthcare facilities can optimize their energy consumption, reduce costs, and enhance their resilience against energy supply fluctuations.

Examining time dynamics and the operational status of healthcare facilities reveals layers of complexity. Energy consumption doesn't remain static throughout the day, with research indicating significant variations influenced by periods of healthcare activity and even seasons. For instance, the interplay between the dynamics of peak hours during the day and seasonal variations such as summer or winter months manifests in unique consumption patterns. This interrelation is further accentuated when considering the operational status of central air conditioning systems, whose activation or deactivation based on certain outdoor temperature thresholds introduces a non-linear relation with temperature. The depth of understanding regarding operational status is currently limited, with gaps evident in literature, thus urging further exploration.

Additionally, as we delve into the domain of machine learning, the crucial role of data preprocessing and feature engineering becomes evident. These processes hold the key to the efficacy of predictive models. From addressing missing values to scaling and normalization, every step-in data preparation has its own set of complexities and considerations. It is imperative to realize that different datasets might demand unique approaches, whether it is filling gaps with preceding measurements, interpolation, or discarding data points altogether. Furthermore, feature engineering holds its own weight. Through it, we seek to improve a model's ability to discern patterns, thereby enhancing prediction accuracy. Techniques like transforming hourly patterns using sine and cosine functions exemplify the finesse involved in this process, revealing cyclic behaviors inherent to energy consumption. Yet, this domain isn't free from challenges. A notable limitation is that a significant portion of research fails to provide detailed methodologies for preprocessing, thereby complicating replication and comparison of results.

As the complexity of energy systems continues to grow, so does the intricacy of predictive modeling. Machine learning models and AI techniques are leading the way in finding innovative solutions in the energy field. The review of literature in this paper revealed three key future research areas in this domain: ANN optimization, the development of new energy optimization techniques and algorithms, and methods to interpret complex models.

**ANN Optimization:** The effectiveness of deep learning models, particularly ANNs in interpreting complex, nonlinear relationships in big datasets is well-established (Wang, et al., 2020). While the capabilities of these structures have been partially explored, much remains to be discovered. A key area of ongoing discussion is determining the optimal configuration of ANNs. Future research could benefit from a methodical investigation into the ideal number of hidden layers and its relationship with a dataset's complexity and size. Additionally, it's essential to comprehend how different activation functions, dropout rates, and other hyperparameters affect the efficiency of the model.

**New Optimization Methods and Algorithms:** The core of any machine learning model is its optimization technique. The path from a model's initial set-up to achieving optimal performance heavily relies on the chosen optimization algorithms. Particularly in energy forecasting through predictive modeling, the choice of optimization algorithm has a significant impact on both the accuracy and computational efficiency. Future investigations into novel optimization methods, such as incorporating evolutionary algorithms or methods based on swarm intelligence, could improve the effectiveness and resilience of predictive models. The exploration of hybrid optimization strategies, which merge the strengths of various algorithms, presents a fruitful direction for achieving comprehensive optimization and avoiding challenges like local minima.

**Complex Model Interpretation:** The shift towards more sophisticated models frequently results in a compromise on interpretability. The unclear characteristics of many advanced models, often referred to as "black boxes," pose challenges, particularly when stakeholders need clear understanding of the decision-making process. Consequently, future research could emphasize enhancing model interpretability, developing methods that offer more straightforward explanations of predictions. This approach could balance the complexity of high-level computational models with the need for understandable results, facilitating more knowledgeable decision-making processes.

While many approaches have been tried and tested within the confines of hospitals, few extend their applications beyond, to other building types. This limited scope hampers our understanding of the generalizability of these AI techniques. Furthermore, a significant portion of studies seems to prioritize short-term predictions. Long-term energy forecasting is also pivotal for strategic planning, remains relatively untouched. Delving deeper into this could unveil how external factors such as global energy policies, technological advancements, and even weather patterns can be integrated for more holistic predictions. Another gap is in understanding potential energy savings. None of the studies have ventured into quantifying the energy savings achievable through precise predictions. The dynamic nature of hospitals, with their ever-evolving patient and staff demographics, mandates the integration of real-time data for predictions. However, this integration, particularly into Intelligent Energy Management Systems (IEMS), remains under-researched. As hospitals become more central in our fight against climate change, ensuring they operate efficiently becomes crucial. The avenues of research identified above will be critical to ensuring that hospitals deliver increasingly energy-efficient, comfortable, and healthy environments for patient care.

## 5 Conclusions

The endeavor to forecast energy consumption in healthcare settings is both critical and complex. While the current body of research offers valuable insights into the factors influencing energy consumption in healthcare buildings, there are significant gaps to be explored. The interplay between occupancy types, their specific activities, and a broad spectrum of meteorological factors complicates the energy prediction landscape. As AI and machine learning continue to advance, the push should be towards integrating more granular and diverse data inputs to refine prediction accuracy further. Only by doing so can we truly harness the potential of AI in optimizing energy consumption in healthcare settings. The relationships between daily operations, seasonal changes, and the specific equipment being used all contribute to the complexity of energy prediction, and they provide interesting avenues for future research. On the other hand, in the world of machine learning, data is absolutely critical. Getting the data ready for analysis, which includes preprocessing and feature engineering, is at the heart of creating models that can make accurate predictions. While there is a lot of existing research on this topic, there are still gaps in our understanding. This underscores the necessity of maintaining transparency in energy forecasting methodologies, pursuing additional research, and engaging in methodical comparisons across various techniques.

The evolution of energy consumption forecasting has been significantly influenced by the advent and refinement of machine learning techniques. Traditional methods like MLR, though foundational, have been surpassed in performance by advanced algorithms such as ANNs. The balance between accuracy and computational speed remains a critical consideration, with models like SVM offering unparalleled precision but at the cost of time. Ensemble methods and hybrid models like WIO-SVR represent innovative approaches, combining the strengths of multiple techniques. Parameter optimization and algorithm refinement also hold promise in enhancing forecasting accuracy. As the energy sector grapples with growing complexity, the quest continues for models that effectively combine accuracy, speed, and interpretability, ensuring both sustainable energy practices and optimal operational efficiency in critical sectors.

# 6  Acknowledgments

This research was supported by H. H. Angus and Associates, Ltd., and the Natural Science and Engineering Research Council [ALLRP-580958-22 and CREATE 510284-2018].

# 7 References

Abdel-Aal & Radwan, E., 2004. Short-term hourly load forecasting using abductive networks. *IEEE Transactions on Power Systems,* pp. 164-173.

Amasyali, K. & El-Gohary, N. M., 2018. A review of data-driven building energy consumption prediction studies. *Renewable and Sustainable Energy Reviews,* Volume 81, pp. 1192-1205.

Bagnasco, A. et al., 2015. Electrical consumption forecasting in hospital facilities: An application case. *Energy and Buildings,* pp. 103, 261–270.

Bekteshi, S. et al., 2015. Dynamic modeling of Kosovo's electricity supply–demand, gaseous emissions and air pollution. *Journal of Sustainable Development of Energy, Water and Environment Systems,* pp. 303-314.

Bui, et al., 2020. Enhancing building energy efficiency by adaptive façade: A computational optimization approach. *Applied energy,* Volume 114797, p. 265.

Buluş, K. & Zor, K., 2021. A hybrid deep learning algorithm for short-term electric load forecasting. *In 2021 29th Signal Processing and Communications Applications Conference,* pp. 1-4.

Cao, L. et al., 2020. Electrical load prediction of healthcare buildings through single and ensemble learning. *Energy Reports.*

Cao, L. et al., 2020. Electrical load prediction of healthcare buildings through single and ensemble learning. *Energy Reports,* p. 2751–2767.

Chalapathy, R., Khoa, N. L. D. & Sethuvenkatraman, S., 2021. Comparing multi-step ahead building cooling load prediction using shallow machine learning and deep learning models. *Sustainable Energy, Grids and Networks,* p. 100543.

Coccagna, M. et al., 2017. Energy consumption in hospital buildings: functional and morphological evaluations of six case studies. *International Journal of Environmental Science,* p. 2.

Dagdougui, H., Bagheri, F., Le, H. & Dessaint, L., 2019. Neural network model for short-term and very-short-term load forecasting in district buildings. *Energy and Buildings,* p. 203.

Fernández-Martínez, D. & Jaramillo-Morán, M. A., 2022. Multi-Step Hourly Power Consumption Forecasting in a Healthcare Building with Recurrent Neural Networks and Empirical Mode Decomposition. *Sensors,* pp. 22(10), 3664.

Hong, T. & Fan, S., 2016. Probabilistic Electric Load Forecasting: A Tutorial Review. *International Journal of Forecasting,* pp. 914-938.

Maddalena, E. T. et al., 2022. Experimental data-driven model predictive control of a hospital HVAC system during regular use. *Energy and Buildings,* pp. 271, 112316.

Manno, A., Martelli, E. & Amaldi, E., 2022. A Shallow Neural Network Approach for the Short-Term Forecast of Hourly Energy Consumption. *Energies,* pp. 15(3), 958.

Nakai, M., Ooka, R. & Ikeda, S., 2021. Study of power demand forecasting of a hospital by ensemble machine learning. *Journal of Physics: Conference Series,* pp. 2069(1), 012147.

Ngo, N.-T.et al., 2021. Proposing a hybrid metaheuristic optimization algorithm and machine learning model for energy use forecast in non-residential buildings. *Scientific Reports,* 12(1).

Olu-Ajayi, R. et al., 2022. Building energy consumption prediction for residential buildings using deep learning and other machine learning techniques. *Journal of Building Engineering,* pp. 45, 103406.

Ozaki, S., Ooka, R. & Ikeda, S., 2021. Energy demand prediction with machine learning supported by auto-tuning: a case study. *Journal of Physics: Conference Series,* pp. 2069(1), 012143.

Panagiotou, D. K. & Dounis, A. I., 2022. Comparison of Hospital Building's Energy Consumption Prediction Using Artificial Neural Networks, ANFIS, and LSTM Network. *Energies,* pp. 54-64.

Raza, M. Q. & Khosravi, A., 2015. A review on artificial intelligence based load demand forecasting techniques for smart grid and buildings. *Renewable and Sustainable Energy Reviews 50,* pp. 352-1372.

Ruiz, E., Pacheco-Torres, R. & Casillas, J., 2017. Energy consumption modeling by machine learning from daily activity metering in a hospital. *n 2017 22nd IEEE International Conference on Emerging Technologies and Factory Automation,* pp. 1-7.

Sarkis-Onofre, R., Catalá-López, F., Aromataris, E. & Lockwood, 2021. How to properly use the PRISMA Statement. *Systematic Reviews,* pp. 10(1), 1-3.

Shao, et al., 2020. Prediction of energy consumption in hotel buildings via support vector machine. *Sustainable Cities and Society 57 ,* p. 102128.

Shiyu, Y. & Wan, M. P., 2022. Machine-learning-based model predictive control with instantaneous linearization–A case study on an air-conditioning and mechanical ventilation system. *Applied Energy,* pp. 306, 118041.

Timur, O. et al., 2020. Application of Statistical and Artificial Intelligence Techniques for Medium-Term Electrical Energy Forecasting: A Case Study for a Regional Hospital. *Journal of Sustainable Development of Energy, Water and Environment Systems.*

Wang, W. et al., 2019. Forecasting district-scale energy dynamics through integrating building network and long short-term memory learning algorithm. *Applied Energy,* pp. 248, 217-230.

Wang, X., Zhao, Y. & Pourpanah, F., 2020. Recent advances in deep learning. *International Journal of Machine Learning and Cybernetics,* 11( ), pp. 747-750.

Wang, Z. & Srinivasan, R. S., 2017. A review of artificial intelligence based building energy use prediction: Contrasting the capabilities of single and ensemble prediction models.. *Renewable and Sustainable Energy Reviews 75 ,* pp. 796-808..

Xue, K. et al., 2020. Investigation and prediction of Energy consumption at St. Olavs Hospital. *In E3S Web of Conferences,* pp. Vol. 246, p. 04003.

Yang, S. & Wan, M. P., 2022. Machine-learning-based model predictive control with instantaneous linearization – A case study on an air-conditioning and mechanical ventilation system. *Applied Energy,* pp. 306, 118041.

Zini, M. & Carcasci, C., 2023. Machine learning-based monitoring method for the electricity consumption of a healthcare facility in Italy. *Energy,* pp. 262, 125576.

Zor, K., Çelik, Ö., Timur, O. & Teke, A., 2020. Short-Term Building Electrical Energy Consumption Forecasting by Employing Gene Expression Programming and GMDH Networks. *Energies,* pp. 13(5), 1102.